\documentclass[a4paper,twocolumn,11pt]{article}
\usepackage[margin=1in]{geometry}
\usepackage{natbib}

\usepackage{array}

\usepackage{lettrine}
\usepackage{enumitem}

\usepackage{amsmath, amssymb}
\usepackage{url,hyperref,cleveref}
\usepackage{booktabs}
\usepackage{hhline} 
\usepackage{colortbl} 
\usepackage{xcolor} 

\usepackage[T1]{fontenc}
\usepackage[utf8]{inputenc}

\renewcommand{\S}[0]{\mathcal{S}}
\newcommand{\M}[0]{\mathcal{M}}
\newcommand{\B}[0]{\mathcal{B}}

\usepackage{tikz}
\usetikzlibrary{shapes.geometric, positioning, arrows.meta, bending,backgrounds,calc}
\usetikzlibrary{decorations.markings}

\usepackage{titlesec}
\titleformat{\section}
  {\normalfont\bfseries\large} % style (slightly smaller than \Large)
  {}                           % no number
  {0pt}                        % no label separation
  {}                           % before-code

\title{Image, Word and Thought: A More Challenging Language Task\\for the Iterated Learning Model}

\author{
\mbox{}} 
\author{
    Hyoyeon Lee, Seth Bullock \& Conor Houghton\\
    \mbox{}\\
    Faculty of Science and Engineering, University of Bristol, England\\
    conor.houghton@bristol.ac.uk
} 
\date{December 23, 2025}

\begin{document}

\maketitle

\begin{abstract}
The iterated learning model simulates the transmission of language from generation to generation in order to explore how the constraints imposed by language transmission facilitate the emergence of language structure. Despite each modelled language learner starting from a blank slate, the presence of a bottleneck limiting the number of utterances to which the learner is exposed can lead to the emergence of language that lacks ambiguity, is governed by grammatical rules, and is consistent over successive generations, that is, one that is expressive, compositional and stable. The recent introduction of a more computationally tractable and ecologically valid \emph{semi-supervised iterated learning model}, combining supervised and unsupervised learning within an autoencoder architecture, has enabled exploration of language transmission dynamics for much larger meaning-signal spaces. Here, for the first time, the model has been successfully applied to a language learning task involving the communication of much more complex meanings: seven-segment display images. Agents in this model are able to learn and transmit a language that is expressive: distinct codes are employed for all 128 glyphs;
compositional: signal components consistently map to meaning components, and stable: the language does not change from generation to generation.\footnote{This is an extended version of a paper accepted for EvoLang2026, it includes additional details of the numerical experiments.}
\end{abstract}

\section{Introduction}

Human language has a distinctive evolutionary character in that it arises from an interplay between at least three adaptive processes: (i) slow evolutionary adaptation of the biological basis of language, (ii) fast individual adaptation through social learning that facilitates language transmission, and (iii) a ``world 3'' intermediate level of cultural-linguistic evolution characterising the adaptation that languages themselves undergo \citep{Popper1972,PopperEccles1977,Christiansen&Kirby2003}. Understanding the interaction between these parallel adaptive processes is challenging given that the processes take place over very different time scales, some of which stretch over many millennia \citep{Deacon1997,ChristiansenChater2008}, records of past languages are incomplete \citep{Bloomfield1933,KirbyHurford2002}, language is both the object of study and the medium through which we communicate our ideas about it \citep{Tomasello1999}, and current language behaviour is instantiated by a neural substrate that is fearsomely complex and poorly understood \citep{PoeppelHickok2004,ChristiansenChater2008}. As such, the debate about the origins and nature of human language spans a wide range of perspectives, from cultural and interaction-driven accounts to proposals emphasizing the co-evolution of symbolic cognition and pluralistic critiques of syntax-centric explanations \citep{Steels2011,Deacon1997,Boeckx2021,Houghton2025}.

Synthetic, computational approaches to understanding the basic properties of language change are attractive since they offer an additional window on evolutionary linguistic phenomena. In particular, computational agent-based models of language learning can offer insight into the way that language learning constraints might influence, shape or determine language properties, potentially offering explanations for the putative universality of some properties of natural language such as the presence of particular grammatical forms or the co-occurrence of specific language features, by exploring ``language as it could be'' \citep{Kirby2002natural}. Typically in models very simple language properties stand in as a proxy for the richer properties of real language; a common example set, the one used here, is expressivity, compositionality and stability \cite{Kirby2001}. 

The iterated learning model is a model \citep[ILM;][]{KirbyHurford2002} of language transmission and evolution which has been successfully applied to simple artificial language learning tasks in which meanings and signals are typically limited to short bit-strings. Here this model is extended to engage with a more complex task in which agents develop a stable, expressive and compositional language with which to communicate successfully about a space of structured images based on the seven-segment LED displays that are often used on calculators, digital watches and digital clocks. 

To our knowledge, this is the first time that an ILM has been demonstrated on a task this complex. This is possible because the ILM used here includes autoencoder learning. In addition to being more computationally tractable and ecologically valid than previous ILMs, the computational architecture of the \emph{semi-supervised ILM} appears to mimic distinct aspects of real-world learning. Extending the model from bit-strings to glyph images required an additional training phase focused only on image encoding and decoding, echoing the separation between low-level sensory representation learning and higher-level learning operating over those representations.

Learning also includes a cognitive scheme that requires agents to employ their language internally as a kind of ``language of thought''. There is a limit to how far such a simple agent model can be interpreted, but one possible reading is that this linking of language and thought corresponds to an important step in the evolution of cognition, in which the internalization of language enables forms of multi-step or chain-of-thought reasoning that have recently proved important in machine learning \citep{WeiEtAl2022}.

Ultimately, our goal is to understand the relationship between language structure and cognitive processes. Simple agent models such as the ILM are can only be viewed as a sourcesof ideas rather than a literal account. From this perspective, the behaviour of the semi-supervised ILM suggests that autoencoder-based learning may play a specific and important role in supporting the emergence and transmission of structured languages, motivating the construction of richer ILM variants such as the one explored here. Future work should include more detailed agent models and real-world experiments; our hope is that the model presented here represents progress towards that goal.

\section{The Iterated Learning Model}

The ILM is an agent-based simulation designed to explore conditions under which languages tend to become expressive, compositional and stable \citep{Kirby2001,KirbyHurford2002}. An ILM comprises a language-using ``tutor'' agent that instructs an initially na\"ive ``pupil'' agent by exposing it to a finite set of utterances, each pairing a meaning with the signal that expresses it in the tutor's language. After this period of instruction, the pupil becomes a tutor and teaches a new initially na\"ive pupil in the same fashion. A language learning bottleneck is a key feature of the model: since each pupil is only exposed to a subset of the tutor's language during learning, when it becomes a tutor itself it will need to generate signals for meanings that were not part of its learning experience. This demand for generalization is critical to the behaviour of the ILM.

Despite there being many different possible languages, most of which map meanings to signals in arbitrary ways, under the right conditions, languages arise within the ILM that are \emph{expressive}, they lack ambiguity, \emph{compositional}, component parts of a signal consistently encode component parts of a meaning in a rule-governed manner, and \emph{stable}, successive pupils learn the same language despite only ever experiencing a small, random fragment of it during learning. The central insight provided by the ILM is that only compositional languages can remain stable and expressive when transmitted through a learning bottleneck because any meaning-signal mapping that \emph{lacks} compositional grammar cannot be consistently generalised to meanings not provided in the bottleneck set.

There are several versions of the ILM, but a typical example due to \cite{KirbyHurford2002} features a space of simple possible languages that each map from a finite set of meanings, $\M$, to a finite set of signals, $\S$, where each meaning and each signal are represented by a binary vector of length $n$. Each language agent's own language is represented by its decoder map $d:\S\rightarrow\M$ which is a simple neural network mapping each possible signal to the meaning that it is used to convey, and its encoder map $e:\M\rightarrow\S$ which is a look-up table mapping each meaning to the signal that conveys it:
\begin{center}
   \begin{tikzpicture}
  \node (S) at (0,0) {$\S$};
  \node (M) at (4,0) {$\M$};
  \draw[-{Latex[bend]}, bend right] (M) to node[above] {$e$} (S);
  \draw[-{Latex[bend]}, bend right] (S) to node[below] {$d$} (M);
\end{tikzpicture}
\end{center}

During learning, a random subset of meanings $\B\in\M$ is chosen and the tutor pairs each with the signal that conveys the meaning in its language $\{(e(b),b)|b\in\B\}$. This set of meaning-signal pairs is used to train the pupil's decoder. At the end of training, the pupil reverse engineers an encoder that is consistent with its decoder, after which it can become a tutor for a newly created na\"ive pupil. Unfortunately, the reverse engineering, or \textsl{obversion}, step is unrealistic in that there is no equivalent step in natural language learning. Furthermore, it scales very badly with the size of the language, making it computationally expensive to the extent that ILM simulations for languages with a bit length above ten are infeasible.

\section{The Semi-Supervised ILM}

In the semi-supervised ILM \citep{BunyanBullockHoughton2025} the agent has two neural networks, an encoder which maps from meaning to signal, a binary vector, and a decoder, which maps from signal to meaning. By removing the discretization step which converts the output of the encoder to binary, the encoder and decoder can also be chained together as an autoencoder. This autoencoder supports learning and serves to couple the encoder and decoder.

The semi-supervised ILM dispenses with the need for obversion by reformulating language transmission as a combination of \emph{supervised} learning episodes and \emph{semi-supervised} learning episodes. The supervised episodes mimic a tutor pointing to a picture of a child stealing biscuits and saying ``Look, a child is stealing biscuits!''. This helps a pupil train both their decoder mapping and their encoder mapping. The \emph{unsupervised} learning episodes correspond to a pupil considering ``How would I describe that image, and what image does that description convey to me?''. This second process subserves two functions. First, it ensures that the decoder and encoder remain consistent so that meanings are encoded into signals that, when decoded, recover the original meanings. Secondly, it prevents a tendency towards over-generalization inherent in language learning. 

A classic real-world example of the tendency towards over-generalization is provided by the word \textsl{thing}. This has gone from referring to a legislative gathering to a general term for any object or concept. It is only by introducing new words for legislative assemblies, such as \textsl{congress}, \textsl{d\'{a}il} or \textsl{parliament}, that it remains easy to discuss, or indeed think about, meetings of this sort. Over-generalization erodes representation of meaning \citep{Orwell1949} but is a natural consequence of the sort of extension-by-metaphor that is usefully employed in discussing novel concepts. Over-generalization is a particular challenge to ILMs. Without some mechanism to prevent it the language quickly collapses towards one without expressivity. By modelling internal reflection where meaning must be preserved to allow the input to be reconstructed, the semi-supervised ILM avoids this danger. It should be noted, however, that the ILM is not a model of communication and this may also play a role in preventing over-generalization.

Each supervised learning episode involves the presentation of a random meaning-signal pair from the tutor's language to either the pupil's encoder network or to their decoder network, using back-propagation to improve incrementally the ability of each network to map the meaning to the signal or the signal to the meaning, respectively. This allows the pupil to learn the language examples provided by the tutor; these are only a partial subset of all possible language examples and the set of examples a tutor teaches do not match the set that tutor was taught. By contrast, unsupervised learning involves the presentation of a random meaning to the input and output layer of the autoencoder network formed by concatenating the pupil's encoder and decoder networks together. This ILM variant has successfully been used to explore ILM behaviour for bit-string languages one thousand times larger than those typically employed before now \citep{BunyanBullockHoughton2025}, and has been used to begin to explore the impact on language structure of phenomena such as language contact \citep{Bullock&Houghton2024}.

%fig1 - this number is the folder in the code github, not the location in the paper
\begin{figure}[t]%bh]
    \centering
    \begin{tabular}{lll}
    \includegraphics[width=0.125\textwidth]{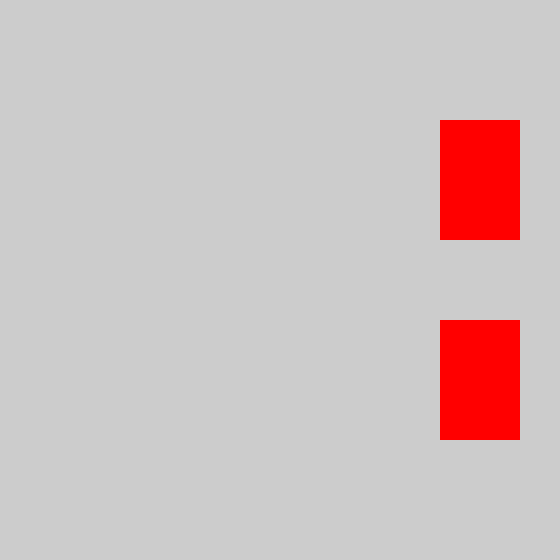}&
    \includegraphics[width=0.125\textwidth]{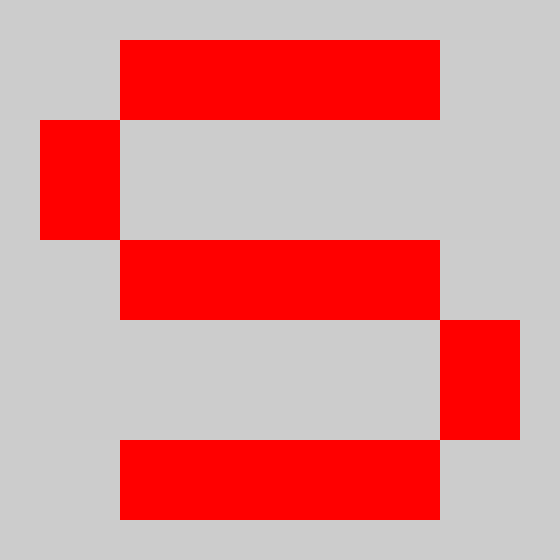}&
    \includegraphics[width=0.125\textwidth]{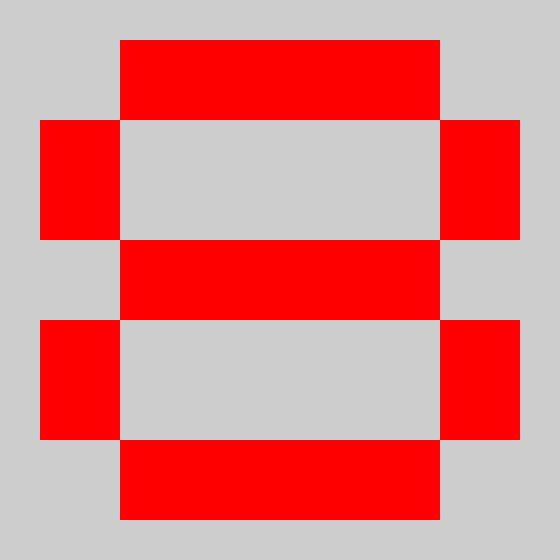}\\
    \end{tabular}
    \caption{\textbf{Example base glyphs.} Each base glyph is a $28\times 28$ image based on the classic seven-segment LED display. All 128 glyphs are employed, many of which do not correspond to digits or letters. The three glyph types presented here correspond to digits 1, 5, and 8. Images are presented as flattened 784-vectors with components between zero and one. The colours used here are for the purpose of illustration, selected to resemble a digital clock display.}
    \label{fig:stimuli}
\end{figure}

\section{The Seven-segment Language Task}

The use of seven-segment displays has a surprisingly long history; in the 1920s they were employed in telephone exchanges that mixed manual and automatic routing \citep{Clark1929} and they became widespread when digital watches and calculators became common. The subsequent introduction of more sophisticated display technologies means that they are not as ubiquitous as they once were. However, it probably remains unnecessary to explain that a seven-segment display is a device used to display digits and some letters by illuminating some combination of seven short vertical and horizontal line segments; three examples are illustrated in Fig.~\ref{fig:stimuli}.  

%fig3
\begin{figure}[tbh]
    \centering
    \begin{tabular}{lll}
    \textbf{A} - \textbf{noise 1}&&\\
    \includegraphics[width=0.125\textwidth]{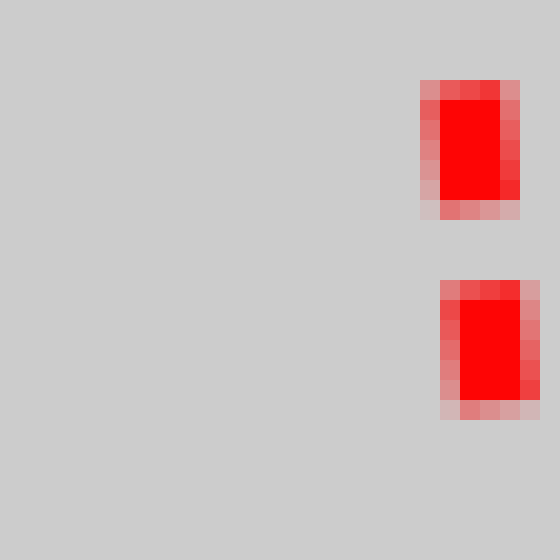}&
    \includegraphics[width=0.125\textwidth]{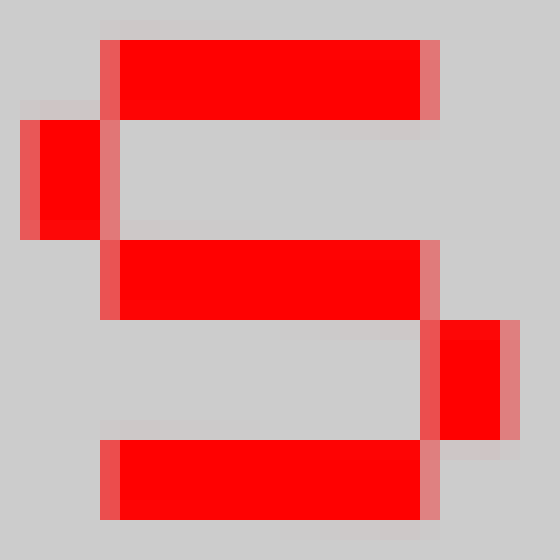}&
    \includegraphics[width=0.125\textwidth]{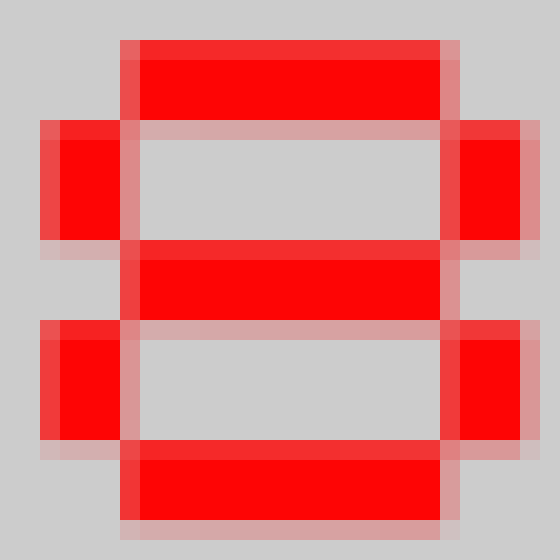}\\
    \textbf{B} - \textbf{noise 2}&&\\
    \includegraphics[width=0.125\textwidth]{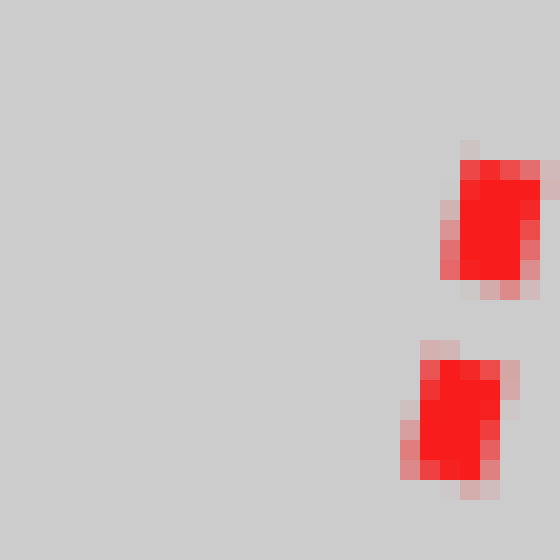}&
    \includegraphics[width=0.125\textwidth]{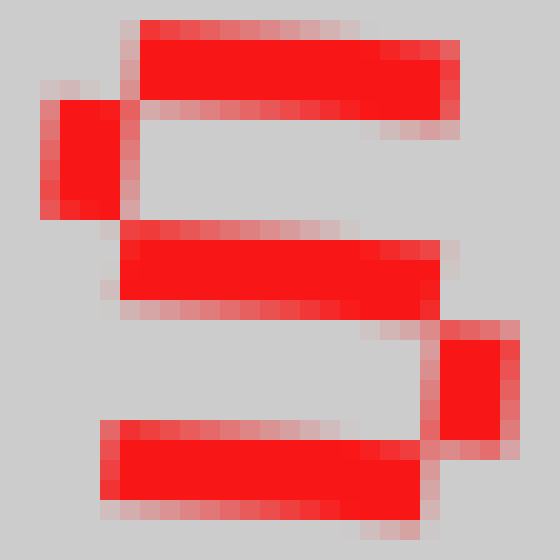}&
    \includegraphics[width=0.125\textwidth]{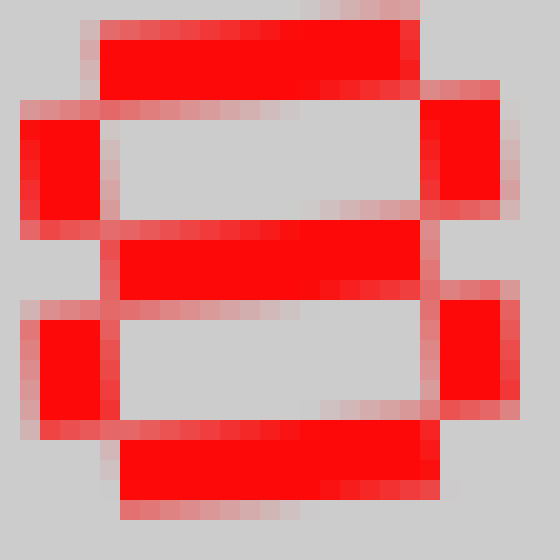}\\
    \textbf{C} - \textbf{noise 3}&&\\
        \includegraphics[width=0.125\textwidth]{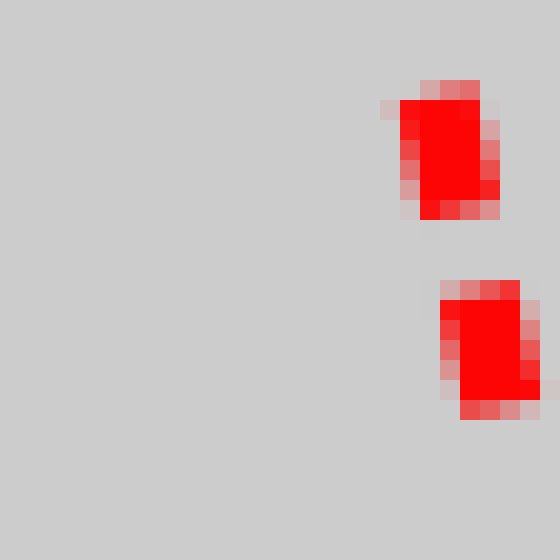}&
    \includegraphics[width=0.125\textwidth]{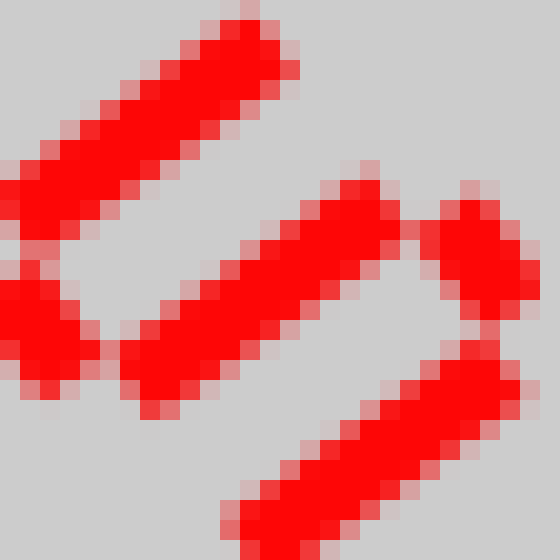}&
    \includegraphics[width=0.125\textwidth]{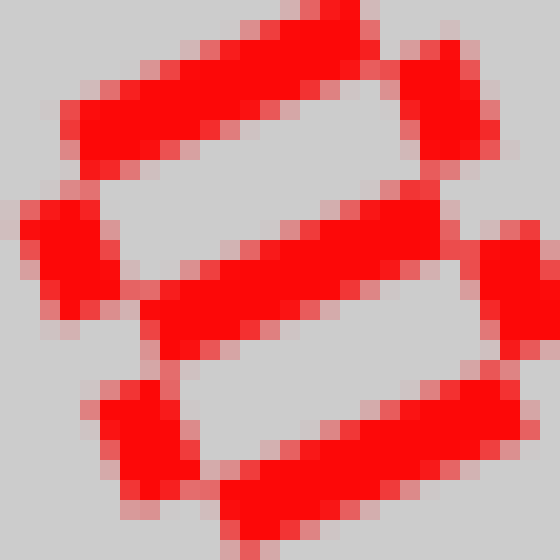}\\
    \end{tabular}
    \caption{\textbf{Example images.} For each base glyph, 100 variant images were produced using small random rotations, translations and changes in intensity. %At the start of each generation, $b=40$ base glyphs are chosen at random and 400 variant images are selected from those associated with the selected base glyphs, meaning that the bottleneck contains, on average, 10 images for each of the selected base glyphs. 
    Example images from each of the three noise conditions described in the text are shown here: \textbf{A}: \textbf{noise 1}, \textbf{B}: \textbf{noise 2} and \textbf{C}: \textbf{noise 3}.}
    \label{fig:images}
\end{figure}

Here, the 128 possible seven-segment display glyphs, including the blank glyph, are employed as a meaning space with compositional structure, that is, glyphs exhibit internal structure. Each base glyph is a $28\times 28$ array with components in $[0,1]$; the dimensions are chosen to match the MNIST dataset. To produce a set of variant images for each of the base glyphs, first each was upsampled by a factor of ten and then three manipulations were used.
\begin{enumerate}
\item Intensity adjustment: the value of the on pixels was changed to $1-\epsilon$ where
\begin{equation}
\epsilon \sim \text{Exponential}(\mu)
\end{equation}
\item Rotation: the image is then rotated by an angle $\theta$ where 
\begin{equation}
\theta\sim \text{Normal}(0,\rho)
\end{equation}
and the angle is measured in radians. 
\item Translation: the image is translated by $\delta x$, $\delta y$ where
\begin{equation}
\delta x, \delta y \sim \text{Normal}(0,\sigma).
\end{equation} 
\end{enumerate}
Finally the image is downsampled back to $28\times 28$. 
Three different levels of noise are considered here:
\begin{enumerate}[labelwidth=4em,leftmargin=!,labelsep=0.5em]
    \item [\textbf{noise 1}:] $\mu=0.03$, $\rho=0.05$, $\sigma=4.0$; Fig.~\ref{fig:images}\textbf{A}
    \item [\textbf{noise 2}:] $\mu=0.1$, $\rho=0.1$, $\sigma=5.0$; Fig.~\ref{fig:images}\textbf{B}
    \item [\textbf{noise 3}:] $\mu=0.1$, $\rho=0.25$, $\sigma=8.0$; Fig.~\ref{fig:images}\textbf{C}
\end{enumerate}
The image dataset, along with the code used to produce it, is available online, see the Data and Code section below.

\section{The Network and the Simulations}

Each language learning agent is represented by an artificial neural network with four parts. When taken together, these constitute an autoencoder configuration, $A$, which can be represented as 

\begin{minipage}{\linewidth}
\vskip 0.25cm
 \centering
    \begin{tikzpicture}[
  node distance=4cm,
  imgDecoder/.style = {
    shape=trapezium,              
    trapezium left angle=116.5650512,          
    trapezium right angle=116.5650512,
    rotate=270,                        % turn it on its side
    draw, fill=blue!20,                % color passed as first arg
    minimum height=1cm,               % this becomes the shorter parallel side (L)
    minimum width=3cm,                % this becomes the longer parallel side (R)
    align=center, font=\small
  },
  imgEncoder/.style = {
    shape=trapezium,              
    trapezium left angle=116.5650512,       
    trapezium right angle=116.5650512,
    rotate=90,                        % turn it on its side
    draw, fill=yellow!20,                % color passed as first arg
    minimum height=1cm,               % this becomes the shorter parallel side (L)
    minimum width=3cm,                % this becomes the longer parallel side (R)
    align=center, font=\small
  },
  wrdEncoder/.style = {
    shape=trapezium,
    trapezium left angle=90,          % make the two non-parallel sides slanted
    trapezium right angle=90,
    rotate=90,                        % turn it on its side
    draw, fill=yellow!20,                % color passed as first arg
    minimum height=1cm,               % this becomes the shorter parallel side (L)
    minimum width=1.5cm,                % this becomes the longer parallel side (R)
    align=center, font=\small
  },  
  wrdDecoder/.style = {
    shape=trapezium,
    trapezium left angle=90,          % make the two non-parallel sides slanted
    trapezium right angle=90,
    rotate=90,                        % turn it on its side
    draw, fill=blue!20,                % color passed as first arg
    minimum height=1cm,               % this becomes the shorter parallel side (L)
    minimum width=1.5cm,                % this becomes the longer parallel side (R)
    align=center, font=\small
  }
]

  %   #1: fill color,    #2: left-side length,  #3: right-side length
  \node[imgEncoder] (imgE) {};
  \node[wrdEncoder,below right  = 1.5cm  and 0cm of imgE] (wrdE) {};
  \node[wrdDecoder,below right=  1.5cm and 0.0cm of wrdE] (wrdD) {};
 \node[imgDecoder,below right= 0.0cm and 1.5cm of wrdD] (imgD) {};
  \node[below left = 1.1cm and 0.5cm of imgE] (imgEl) {$E_i$};
  \node[below left = 1.1cm and 0.5cm of wrdE] (wrdEl) {$E_w$};
  \node[below left = 1.1cm and 0.5cm of wrdD] (wrdDl) {$D_w$};
  \node[below right = 1.1cm and 0.5cm of imgD] (imgDl) {$D_i$};
\end{tikzpicture}

\end{minipage}
The first section $E_i$ is the \emph{image encoder}, this is a $784\times 128\times n_l$ feedforward all-to-all neural network. The size of the first layer is, of course, fixed by the dimension of the images. The middle layer dimension could vary, but the same value is used for all the simulations reported here. The latent dimension, $n_l$, is the size of the output layer of $E_i$ and of the input layer of $E_w$, the \emph{word encoder}. This is a $n_l\times n_l\times n_l$ feedforward all-to-all neural network. Next, $D_w$, the \emph{word decoder}, is another $n_l\times n_l\times n_l$ feedforward all-to-all neural network. Finally the \emph{image decoder}, $D_i$, is a $n_1\times 128\times 784$ feedforward all-to-all neural network. For most of the simulations reported here $n_l=7$ or $n_l=10$; though in one case the word encoder and word decoder have different sizes for different layers. As a useful convention, networks will be referred to by a capital letter, the corresponding map by a small letter.

An agent's network is not trained in the specific autoencoder configuration in $A$; in fact there are four different combinations that are used in training. An agent's encoder is $E=\delta\circ E_w\circ E_i$:

\begin{minipage}{\linewidth}
 \centering
   \begin{tikzpicture}[
  node distance=4cm,
  imgEncoder/.style = {
    shape=trapezium,              
    trapezium left angle=116.5650512,          % make the two non-parallel sides slanted
    trapezium right angle=116.5650512,
    rotate=90,                        % turn it on its side
    draw, fill=yellow!20,                % color passed as first arg
    minimum height=1cm,               % this becomes the shorter parallel side (L)
    minimum width=3cm,                % this becomes the longer parallel side (R)
    align=center, font=\small
  },
  wrdEncoder/.style = {
    shape=trapezium,
    trapezium left angle=90,          % make the two non-parallel sides slanted
    trapezium right angle=90,
    rotate=90,                        % turn it on its side
    draw, fill=yellow!20,                % color passed as first arg
    minimum height=1cm,               % this becomes the shorter parallel side (L)
    minimum width=1.5cm,                % this becomes the longer parallel side (R)
    align=center, font=\small
  }
]

  %   #1: fill color,    #2: left-side length,  #3: right-side length
  \node[imgEncoder] (imgE) {};
  \node[wrdEncoder,below right  = 1.5cm  and 0cm of imgE] (wrdE) {};
    \node[below right= 0.0cm and 0cm of wrdE,
    draw,
    rectangle,
    fill=lightgray,
    minimum width=0.5cm,
    minimum height=1.5cm,
    inner sep=0pt
  ] (delta) {};

\end{tikzpicture}
 
\end{minipage}
The map $\delta$ is the discretization map $\delta z\rightarrow (z > 0.5\,?\, 1 : 0)$
and so $E$ maps images to binary $n_l$-vectors, what are regarded here as \emph{words}. An agent's decoder is $D=D_i\circ D_w$:

\begin{minipage}{\linewidth}
 \centering
   \begin{tikzpicture}[
  node distance=4cm,
  imgDecoder/.style = {
    shape=trapezium,              
    trapezium left angle=116.5650512,          % make the two non-parallel sides slanted
    trapezium right angle=116.5650512,
    rotate=270,                        % turn it on its side
    draw, fill=blue!20,                % color passed as first arg
    minimum height=1cm,               % this becomes the shorter parallel side (L)
    minimum width=3cm,                % this becomes the longer parallel side (R)
    align=center, font=\small
  },
  imgEncoder/.style = {
    shape=trapezium,              
    trapezium left angle=116.5650512,          % make the two non-parallel sides slanted
    trapezium right angle=116.5650512,
    rotate=90,                        % turn it on its side
    draw, fill=yellow!20,                % color passed as first arg
    minimum height=1cm,               % this becomes the shorter parallel side (L)
    minimum width=3cm,                % this becomes the longer parallel side (R)
    align=center, font=\small
  },
  wrdEncoder/.style = {
    shape=trapezium,
    trapezium left angle=90,          % make the two non-parallel sides slanted
    trapezium right angle=90,
    rotate=90,                        % turn it on its side
    draw, fill=yellow!20,                % color passed as first arg
    minimum height=1cm,               % this becomes the shorter parallel side (L)
    minimum width=1.5cm,                % this becomes the longer parallel side (R)
    align=center, font=\small
  },  
  delta/.style = {
    shape=trapezium,
    trapezium left angle=90,          % make the two non-parallel sides slanted
    trapezium right angle=90,
    rotate=90,                        % turn it on its side
    draw, fill=black!20,                % color passed as first arg
    minimum height=0.5cm,               % this becomes the shorter parallel side (L)
    minimum width=1cm,                % this becomes the longer parallel side (R)
    align=center, font=\small
  },
    wrdDecoder/.style = {
    shape=trapezium,
    trapezium left angle=90,          % make the two non-parallel sides slanted
    trapezium right angle=90,
    rotate=90,                        % turn it on its side
    draw, fill=blue!20,                % color passed as first arg
    minimum height=1cm,               % this becomes the shorter parallel side (L)
    minimum width=1.5cm,                % this becomes the longer parallel side (R)
    align=center, font=\small
  }
]

  %   #1: fill color,    #2: left-side length,  #3: right-side length
  \node[wrdDecoder] (wrdD) {};
 \node[imgDecoder,below right= 0cm and 1.5cm of wrdD] (imgD) {};
\end{tikzpicture}
 
\end{minipage}
which maps words to images. 

In the semi-supervised ILM, the network is also trained as an autoencoder. Here, this means training two different autoencoder configurations. The \textsl{outer} autoencoder, $O=D_i\circ E_i$:

\begin{minipage}{\linewidth}
\vskip 0.25cm
 \centering
\begin{tikzpicture}[
  node distance=4cm,
  imgDecoder/.style = {
    shape=trapezium,              
    trapezium left angle=116.5650512,          % make the two non-parallel sides slanted
    trapezium right angle=116.5650512,
    rotate=270,                        % turn it on its side
    draw, fill=blue!20,                % color passed as first arg
    minimum height=1cm,               % this becomes the shorter parallel side (L)
    minimum width=3cm,                % this becomes the longer parallel side (R)
    align=center, font=\small
  },
  imgEncoder/.style = {
    shape=trapezium,              
    trapezium left angle=116.5650512,          % make the two non-parallel sides slanted
    trapezium right angle=116.5650512,
    rotate=90,                        % turn it on its side
    draw, fill=yellow!20,                % color passed as first arg
    minimum height=1cm,               % this becomes the shorter parallel side (L)
    minimum width=3cm,                % this becomes the longer parallel side (R)
    align=center, font=\small
  }
]

  %   #1: fill color,    #2: left-side length,  #3: right-side length
  \node[imgEncoder] (imgE) {};
 \node[imgDecoder,below right= 0cm and 1.5cm of imgE] (imgD) {};
\end{tikzpicture}

\end{minipage}
is intended to stand in for the automatic development of the primary visual system, the sort of refinement that occurs through neural processes within the primary sensory systems. An example of this is the activity‐driven refinement of retinotopic maps in the mammalian visual system: before the eyes even open, spontaneous retinal waves propagate across the neonatal retina and without any higher‐order feedback, drive the sharpening of both  projection fields and ocular‐dominance columns \citep{MeisterEtAl1991,ShatzStryker1978,CangFeldheim2013}.

The inner autoencoder,

\begin{minipage}{\linewidth}
 \centering
\begin{center}
    \begin{tikzpicture}[
  node distance=4cm,
 wrdEncoder/.style = {
    shape=trapezium,
    trapezium left angle=90,          % make the two non-parallel sides slanted
    trapezium right angle=90,
    rotate=90,                        % turn it on its side
    draw, fill=yellow!20,                % color passed as first arg
    minimum height=1cm,               % this becomes the shorter parallel side (L)
    minimum width=1.5cm,                % this becomes the longer parallel side (R)
    align=center, font=\small
  },  
  wrdDecoder/.style = {
    shape=trapezium,
    trapezium left angle=90,          % make the two non-parallel sides slanted
    trapezium right angle=90,
    rotate=90,                        % turn it on its side
    draw, fill=blue!20,                % color passed as first arg
    minimum height=1cm,               % this becomes the shorter parallel side (L)
    minimum width=1.5cm,                % this becomes the longer parallel side (R)
    align=center, font=\small
  }
]

  %   #1: fill color,    #2: left-side length,  #3: right-side length
  \node[wrdEncoder] (wrdE) {};
  \node[wrdDecoder,below right=  1.5cm and 0.0cm of wrdE] (wrdD) {};

\end{tikzpicture}

\end{center}
\end{minipage}
is intended to mimic internal reflection; the output of the visual system is mapped by $E_w$ to the latent layer, the layer that produces signals, and then is decoded back by $D_w$ to recover the original signal. The model requires an inner autoencoder if it is to develop expressiveness; without it the dynamics of the model pushes the language towards a smaller and smaller set of words. This second autoencoder can be thought of as modelling internal cognitive reflection, analogous to the internal rehearsal or inner speech processes hypothesised to play a critical role in language learning and cognitive development \citep{Vygotsky1986, Carruthers2002}.

At the start of each pupil's education, $b=40$ glyphs are selected at random from the full set of 128; 400 images are then picked randomly from the full set of images corresponding to those 40 glyphs. The tutor's encoder, $E$, is used to construct the bottleneck set of 400 image-signal pairs. An additional ``everyday'' set of 1200 images are chosen at random from the full set of $128\times100$ variant images. 

A training epoch has 400 iterations, one for each element in the bottleneck set; in each iteration a pair from $\cal{B}$ is used to train the pupil's encoder $E$ and a pair is used to train the pupil's decoder $D$; $r=15$ images are then selected from the everyday set to train the outer encoder, $O$, and another $r=15$ are used to train the inner encoder, $I$; the pupil's own image encoder is used to convert these images into input for $I$. All training used a stochastic gradient descent with a learning rate of $\eta=15$ and $n_e=15$ epochs of training for each pupil; the loss function is MSE-loss; all non-linearities are sigmoids.

\section{Evaluation}

%fig2
%\begin{figure}[tbh]
%    \centering
%    %\begin{tabular}{lll}
%    %\textbf{A}: expressivity&\textbf{B}: compositionality&\textbf{C}: stability\\
%    \textbf{A}: expressivity\\%&\textbf{B}: compositionality&\textbf{C}: stability\\
%    \includegraphics[width=0.35\textwidth]{figures/fig2_express.pdf}\\
%    \textbf{B}: compositionality\\%&\textbf{C}: stability\\
%    \includegraphics[width=0.35\textwidth]{figures/fig2_compose.pdf}\\
%    \textbf{C}: stability\\
%    \includegraphics[width=0.35\textwidth]{figures/fig2_stable.pdf}\\
%    %\end{tabular}
%    \caption{\textbf{An XCS language evolves.} Here the expressivity, $x$, compositionality, $c$, and stability, $s$, of the pupil's encoder at the completion of training are plotted over 100 generations for a $784\times 128\times 7^5\times 128\times 784$ model. All three properties approach one. In all cases, the main, thicker line shows the average of ten instantiations while the thin lines correspond to individual instantiations.}
%    \label{fig:xcs}
%\end{figure}

\begin{figure*}[tbh]
    \centering
    \begin{tabular}{lll}
    \textbf{A}: expressivity&\textbf{B}: compositionality&\textbf{C}: stability\\
    %\textbf{A}: expressivity\\%&\textbf{B}: compositionality&\textbf{C}: stability\\
    \includegraphics[width=0.3\textwidth]{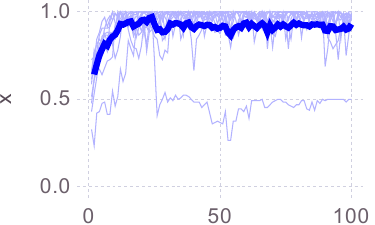}&
    %\textbf{B}: compositionality\\%&\textbf{C}: stability\\
    \includegraphics[width=0.3\textwidth]{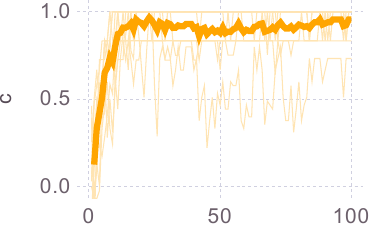}&
    %\textbf{C}: stability\\
    \includegraphics[width=0.3\textwidth]{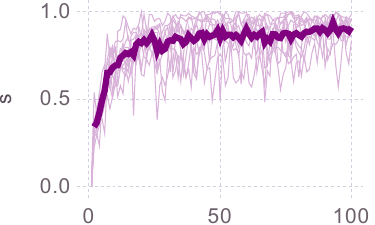}\\
$\qquad\qquad\qquad\quad$generations&$\qquad\qquad\qquad\quad$generations&$\qquad\qquad\qquad\quad$generations
    \end{tabular}
    \caption{\textbf{An expressive, compositional and stable language evolves.} Thin lines show expressivity, compositionality and stability for ten individual 100-generation instantiations with $n_l=7$. Thick lines show the average values; all three measures are defined so that they range from zero to one.}
    \label{fig:xcs}
\end{figure*}

The model's language is evaluated against three quantities, $x$, $c$ and $s$, corresponding to expressivity, compositionality and stability. The set of all base glyphs $\mathcal{G}$ ar used to calculate measures of $x$ and $s$. Let $s_i=e(g_i)$ where $g_i\in \mathcal{G}$ and $e$ is the map corresponding to the pupil's encoder at the end of training; the expressivity is the size of this set of signals $x=|\{s_i|i\in 1:128\}|/128$. Language stability is calculated by comparing the signals used by pupils to those used by tutors, using the obvious notation $s=|\{s_i^t==s_i^{t-1}|i\in 1:128\}|/128$. 

Compositionality is always tricky to define. Here, a proxy measure is used. This measure has the advantage of being relatively simple. It uses the seven base glyphs where only a single segment is on which we term the set of \textsl{one-hot glyphs}. In a maximally compositional code each one-hot glyph should be coded for by a single unique bit in the signal. To calculate the measure, the seven signals corresponding to these one-hot glyphs are calculated and composed into a matrix $C=[s_1|s_2|\ldots|s_7]$. If $n_l=7$ this is a square matrix. A single signal bit can code for an individual segment if it is either one for that segment and zero otherwise, or zero for that segment and one otherwise; to account for this each row of $C$ is examined and if it has four or more ones then the values in that row are flipped, with ones changing to zeros and zeros to ones. Next the row sums and the column sums are calculated, $r_a=\sum_i C_{ai}$ and $c_a=\sum_i C_{ia}$. These values are scored using a sort of rectified reciprocal $\phi(r)=(r>0\,?\, 1/r : 0)$ and the compositionality is then 
\begin{equation}
    \tilde{c}=\frac{1}{14}\left(\sum_a [\phi(r_a)+\phi(c_a)]\right)
\end{equation}
When $n_l>7$ this value is calculated for all possible combinations of seven out of $n_l$ columns and $\tilde{c}$ is the largest value. Finally the background value, $c_0$, is calculated by constructing sets of seven $n_l$-vectors with components drawn from Bernouilli(0.5) and used to normalise the calculation of compositionality as $c=(\tilde{c}-c_0)/(1-c_0)$. 

This proxy measure for $c$ is not perfect: while it achieves a maximal value of one for a compositional language, a value of one does not, in principle, guarantee that the language is compositional. For a compositional language each one-hot glyph must be coded for by one bit and each bit must code for only one glyph. This is, indeed, required for $c=1$ using the definition here. However, for a compositional language, the same bit that codes for a particular segment in one glyph must always code for that same segment in every glyph. This is not required for $c=1$ and so the measure could have the value one for a non-compositional language. Visually inspecting the output, for example using the sort of graph in Fig.~\ref{fig:latents}, indicates that this does not happen and the measure appears to work in practice as a proxy for compositionality.

\section{Results}

%fig4
\begin{figure}[tbh]
    \centering
    \begin{tabular}{llll}
    &\textbf{encoder}&\textbf{decoder}&\\
    \rotatebox{90}{$\quad\qquad$loss}&
    \includegraphics[width=0.15\textwidth]{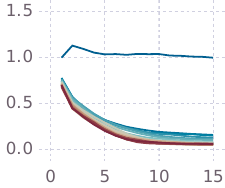}&
    \includegraphics[width=0.15\textwidth]{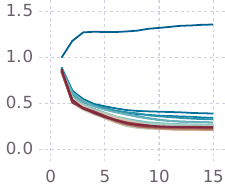}&\\
       &\textbf{inner}&\textbf{outer}\\
    \rotatebox{90}{$\quad\qquad$loss}&
    \includegraphics[width=0.15\textwidth]{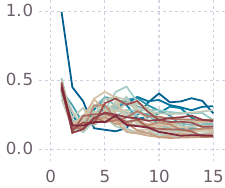}&
    \includegraphics[width=0.15\textwidth]{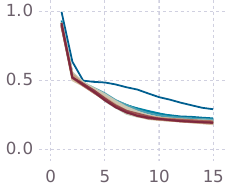}&
    \includegraphics[width=0.05\textwidth]{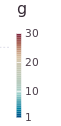}\\
    &$\quad\qquad$~epoch&$\quad\qquad$~epoch
    \end{tabular}
     \caption{\textbf{A compositional and expressive language is easier to learn}. Loss has been plotted against epoch during pupil training. Four different parts of a pupil's network for $n_l=7$ are trained for $g=30$ generations and the loss for each of these is plotted here; different generations distinguished by color and warmer colors corresponding to later generations. The loss for the first epoch of the first pupil is used to normalize the plots and ten instantiations have been averaged.}
    \label{fig:epochs}
\end{figure}

%fig5
\begin{figure}[ht]
    \begin{tabular}{l@{\hspace{0.1em}} lll}
    &\textbf{outer}&\textbf{autoencoder}&\textbf{decoder}\\
     \rotatebox{90}{~~\textbf{6 generations}}&
    \includegraphics[width=0.125\textwidth]{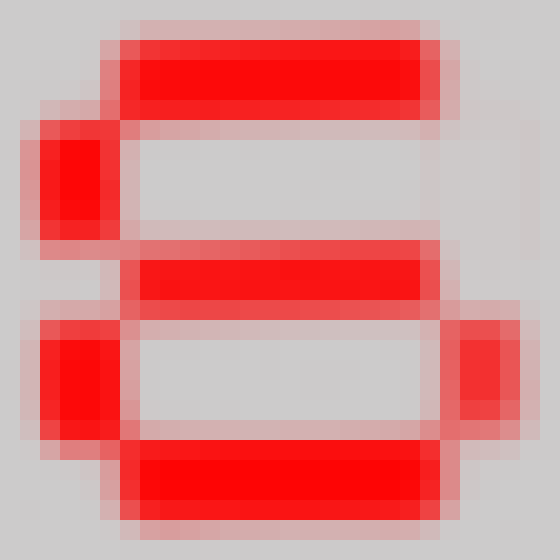}&
    \includegraphics[width=0.125\textwidth]{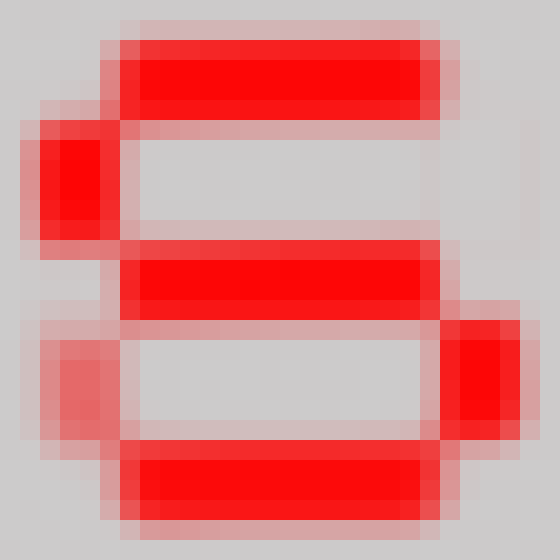}&
    \includegraphics[width=0.125\textwidth]{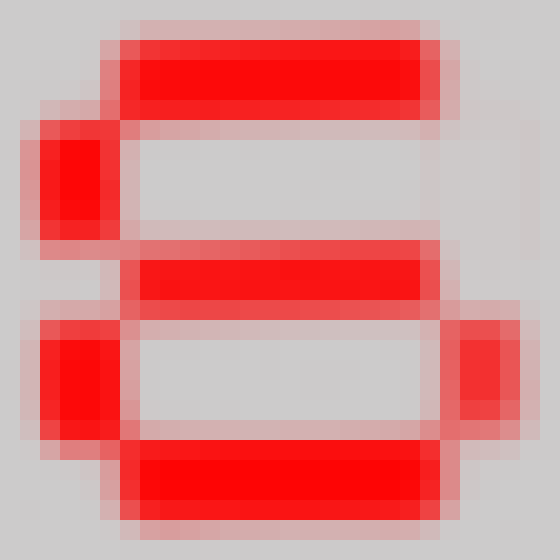}\\[10 pt]
     \rotatebox{90}{\textbf{30 generations}}&
    \includegraphics[width=0.125\textwidth]{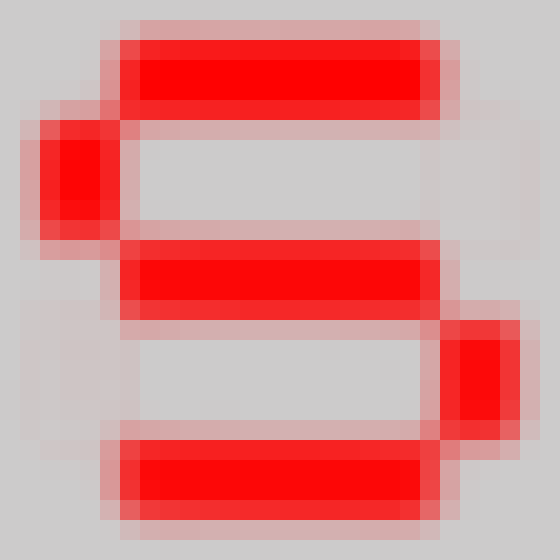}&
    \includegraphics[width=0.125\textwidth]{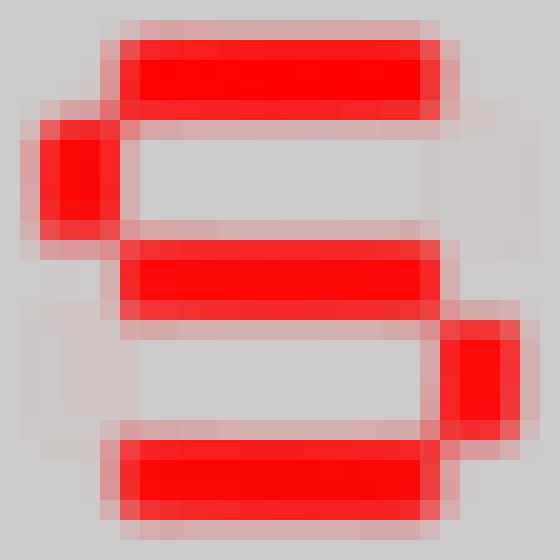}&
    \includegraphics[width=0.125\textwidth]{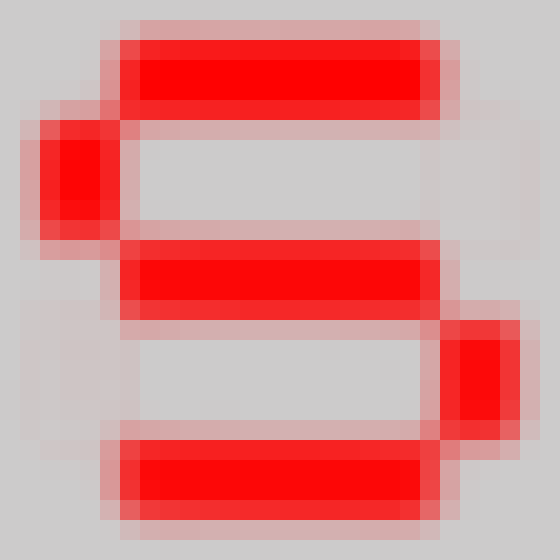}\\    
    \end{tabular}
    \caption{\textbf{Reconstructed images for the 5 glyph}. The top row presents reconstructions after six generations, the bottom row, after 30. Three different types of reconstruction are shown: the first column, \textbf{outer}, shows reconstruction by the outer autoencoder, $O$; the second, \textbf{autoencoder}, shows reconstruction by the total autoencoder, $A$. The third column, \textbf{decoder} is similar, it is $D\circ E$ so it differs from the autoencoder by including the discretization step. After six generations the network is already making compositional mistakes, inclusion or omission of entire segments, but the different types of autoencoding are not fully aligned to each other. By generation 30 the reconstructions are nearly identical. This example used $n_l=10$.}
    \label{fig:reconstruct_five}
\end{figure}

Although this task is more complex than any the ILM has been applied to before now, the model succeeds in developing an expressive, compositional and stable language. This is demonstrated in Fig.~\ref{fig:xcs} which plots the course of evolution for $n_l=7$ using the \textbf{noise 1} image dataset. All of $x$, $c$ and $s$ quickly approach values near one. As the language becomes more compositional, it becomes easier to learn; this is demonstrated in Fig.~\ref{fig:epochs}. Loss falls faster in later generations and reaches a lower value; this effect is most obvious for the inner autoencoder, which is most affected by competition between the supervised and unsupervised learning. The effect is least obvious for the outer autoencoder, $O$.

With just seven bits, a compositional language is unable to code for any aspect of the noise and the actual reconstructions are poor. In Fig.~\ref{fig:reconstruct_five} the reconstructed images for a network with larger, $n_l=10$, words are shown. While the input is a base glyph, the network was trained on the \textbf{noise 1} image dataset so the reconstruction is slightly smeared-out, but clearly successful.

%fig6 / fig8
\begin{figure*}[tbh]
    \centering
    \begin{tabular}{lll}
    \textbf{A}: \textbf{noise 2} seven latent nodes&&\\
    \includegraphics[width=0.3\textwidth]{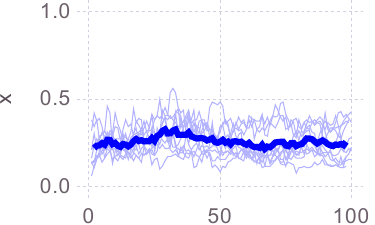}&
    \includegraphics[width=0.3\textwidth]{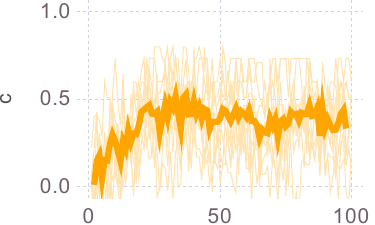}&
    \includegraphics[width=0.3\textwidth]{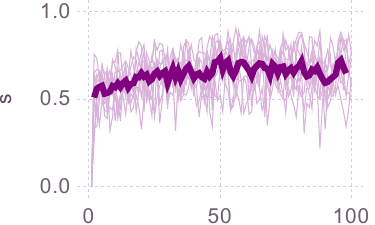}\\
    \textbf{B}: \textbf{noise 2} ten latent nodes &&\\
        \includegraphics[width=0.3\textwidth]{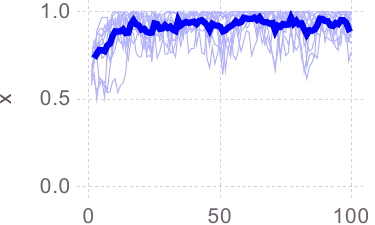}&
    \includegraphics[width=0.3\textwidth]{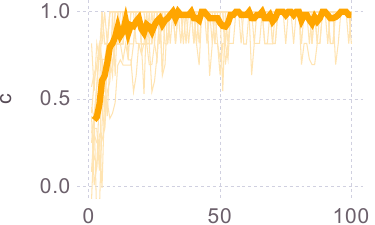}&
    \includegraphics[width=0.3\textwidth]{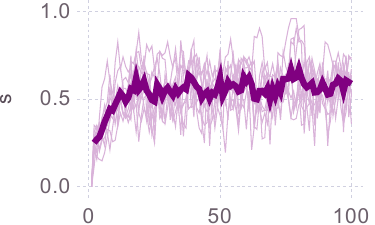}\\
    \textbf{C}: \textbf{noise 3}&&\\
           \includegraphics[width=0.3\textwidth]{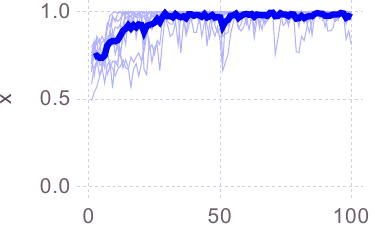}&
    \includegraphics[width=0.3\textwidth]{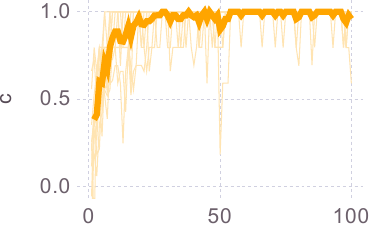}&
    \includegraphics[width=0.3\textwidth]{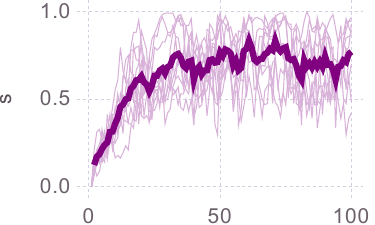}\\ $\qquad\qquad\qquad\quad$generations&$\qquad\qquad\qquad\quad$generations&$\qquad\qquad\qquad\quad$generations
    \end{tabular}
    \caption{\textbf{ILM performance with noisier images}. The columns here follow the same order, $x$, $c$ and $s$ as in Fig.~\ref{fig:xcs}. In the top two rows, \textbf{A} and \textbf{B}, the \textbf{noise 2} images are used. The network with $n_l=7$ plotted in \textbf{A} does not establish a good language. However, expanding the number of latent dimensions so $n_l=10$ allows an expressive and compositional languages to arise. Their stability remains modest. With a large enough number of latent nodes the model can achieve an expressive and compositional language even for \textbf{noise 3} images, \textbf{C} show results where $E_w$ is $20\times 18\times 15$.}
    \label{fig:XCSnoisier}
\end{figure*}

The model also succeeds in evolving an expressive and compositional language for the \textbf{noise 2} dataset; however, this requires a larger value of $n_l$. In Fig.~\ref{fig:XCSnoisier}\textbf{A} a network with $n_l=7$ latent nodes is used with the \textbf{noise 2} dataset and the values of $x$ and $c$ remain largely below 0.5. The stability, $s$, is greater, but high stability alone is not impressive since two agents that use a language that maps every meaning to the same signal will be scored high on stability. While nothing so extreme has happened here, it is clear that the model has failed. However, increasing the number of latent nodes to $n_1=10$ allows an expressive and compositional language to quickly arise, Fig.~\ref{fig:XCSnoisier}\textbf{B}. With a large enough network, an expressive and compositional language can also be achieved for \textbf{noise 3} images, Fig.~\ref{fig:XCSnoisier}\textbf{C}. 

\begin{figure*}[tbh]
\centering
%            col1            gap↑         col2            gap↓         col3   col4
\begin{tabular}{l@{\hspace{0.15em}}lll@{\hspace{0.9em}}}
\textbf{A}&one&five&\\
% col1: vertically centered image

% col2: rotated labels (no internal \quad/\qquad)
\rotatebox{90}{~$\qquad$images}&
% col3, col4: figures
\includegraphics[width=0.3\textwidth]{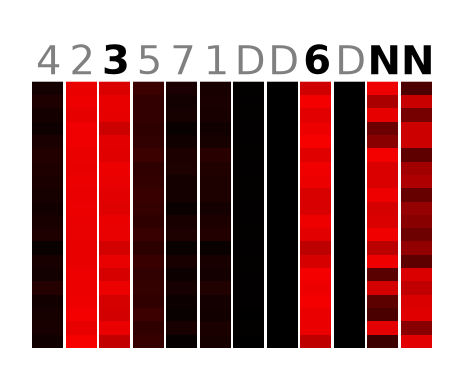}&
\includegraphics[width=0.3\textwidth]{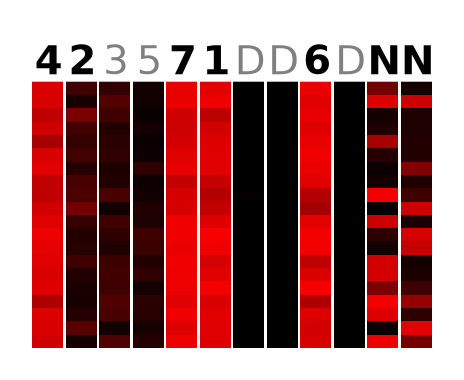}&
\raisebox{0.25\height}{\includegraphics[width=0.15\textwidth]{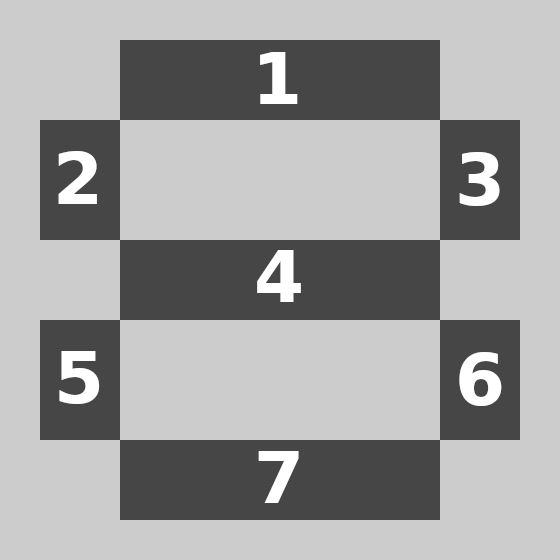}}\\
\textbf{B}&one&five&\\
\rotatebox{90}{$\quad$generations}&
\includegraphics[width=0.3\textwidth]{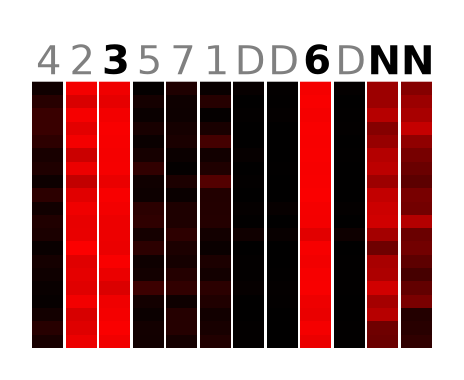}&
\includegraphics[width=0.3\textwidth]{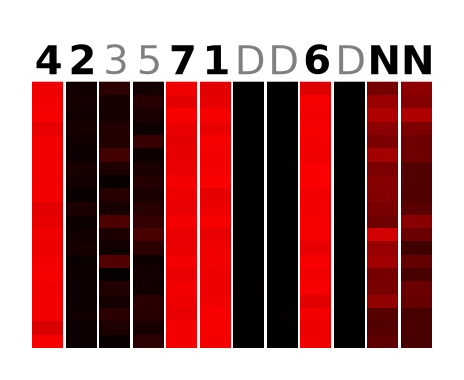}&
\end{tabular}

    \caption{{\footnotesize{}\textbf{Latent activations}. These show latent activations for two glyphs. In every case the columns correspond to latent nodes. Some of the columns have been labelled according to which glyph segment they respond to; the active segments are labelled in bold. For \textbf{A}, generation 60 is considered, at which point $x=1.0$ and $c=1.0$. For each glyph, each row depicts how one of 20 \textbf{noise 2} 1 or 5 images are mapped to latent activations using the encoder $E$. No discretization is applied, though latent activations are frequently close to either zero or one. Many remain the same across variant images; these, presumably, correspond to the compositional coding of glyph segments. For \textbf{B} generations 50 to 69 are considered, for which $s=0.496$. Each row corresponding to the latent values for the base glyph at a different generation, lower rows correspond to later generations. The same columns that show variation here are the ones that show variation across images in \textbf{A}. For these simulations a $16\times 14\times 12$ $E_w$  and the noise 3 data were used; this large latent space helps the network learn quickly, but it appears some of the latent nodes, labelled D, cease to be active. Of the unlabelled columns, only the last two, labelled N, show activity in representing the noise. In this example only the second latent node codes for a glyph segment ``backwards'', with low activation representing the presence of segment 2.}}
    \label{fig:latents}
\end{figure*}

Interestingly, in these examples the language never stabilizes. At first this seems unexpected. The structure of a compositional language means that it generalizes from one bottleneck set to the next; something that makes it easier to learn and something that should facilitate stability. However, when $n_l=10$ for example, compositionality is only seen in seven of the ten signal bits and the compositional structure of the images is accounted for by these bits alone; additional differences between images of the same glyph arise because of noise and it can be imagined that the additional three bits could provide some information about how noise has affected different images of a particular glyph. However, it seems that this ILM is unable to find a compositional structure in the noise and so the additional bits are not stable. This is demonstrated in Fig.~\ref{fig:latents} where it is observed that the bits that vary between different variant images of the same glyph are also the bits that vary from generation to generation. In this case, a shared language has arisen that can be used to communicate the identity of each glyph in a way that is stable across generations despite limited opportunity to learn from previous language users, but while agents also develop the ability to represent the perturbations that these glyphs are subjected to, this feature of the language is stubbornly idiosyncratic. 

%figure 7

\section{Discussion}

The results presented here demonstrate that the semi-supervised ILM can successfully evolve expressive, compositional, and stable languages by discovering the compositional structure in noisy seven-segment display images. This marks an advance in terms of the complexity of the space of meanings. Training for this model involves the obvious supervised training using exemplars supplied by the tutor and the unsupervised autoencoder learning previously introduced in \citep{BunyanBullockHoughton2025}. To succeed in this more complicated case the autoencoder was split into two configurations, an outer autoencoder which allowed the network to learn the primary structure of the data, analogous to how the basic structure of visual data is learned in V1, and an inner autoencoder which learns the higher level compositional structure of the data.

Clearly the results reported here are preliminary: we have not explored the parameter space by varying the size of the bottleneck or, in a systematic fashion, the architecture of the network. More importantly, we have not mapped the contours of compositionality. One of the most interesting aspect of our results is that the network appears able to exploit the compositional structure of the seven-segment images but not the structure of the noise; in future it would be useful to consider image variations in, for example, location or orientation that interpolate between what might be considered image properties and what might be considered noise.

Ultimately our goal is to use the ILM to study the relationship between language structure, language transmission and the influence on language of the structure of thought. In this regard it is useful to have successfully used the ILM with a more demanding input data and it is interesting that this required splitting the training between the outer and inner autoencoder, with the first acting to discover the structure of the input and the second finds a compositional langauge patterned on that structure. This shows the potential of computational models like the semi-supervised ILM to illuminate the interaction between linguistic signals and their cognitive representations. Future research can build upon this foundation by extending the model to even richer representational domains and by exploring the conditions under which additional natural language properties, such as hierarchical syntax or pragmatic inference, might spontaneously emerge.

\section{Conclusion}

Previous ILM studies have explored several important and interesting types of language change phenomena, such as language learning within socially stratified, spatially distributed or linguistically diverse populations \citep{BraceBullock2015,BraceBullock2016ALIFE,SainsEtAl2023,Bullock&Houghton2024}. However, these studies have often been limited to considering simple meaning-signal spaces or hand-designed grammars, neither of which allowed models to explore the emergence of complex natural language features such as hierarchical sentence structure, word order, agreement and inflection. 

In this paper a class of semi-supervised ILM was employed to explore the dynamics of language transmission for a more challenging task than has been previously addressed within the ILM literature. The results reported here demonstrate that, when challenged to communicate about a space of noisy images that exhibit internal, compositional structure, semi-supervised ILM agents are capable of developing their own expressive, compositional language and successfully transmitting it through a language learning bottleneck. This establishes a significant platform upon which to build simulation studies of language change within which the emergence and transmission of complex grammars can be explored and the interplay between the multiple adaptive processes responsible for natural language change dynamics can be explained. 

The ILM is a very simple model but in the semi-supervised form employed here it supports a striking hypothesis about the origin of language: rather than being wholly concerned with encoding and decoding external utterances from a shared language, the agent must also rely on autoencoder-driven learning of its own ``language of thought'' \cite{Vygotsky1986,Carruthers2002}.  Clearly, there is a substantial gap between the simplicity of the model and the subtlety of the language of thought idea; to bridge this the model needs to be applied to more complex examples. This paper is a first step in this programme and comes with an interesting lesson, the same autoencoder mechanism that is useful for language learning and crucial for language evolution is also useful in learning the more peripheral parts of the pathway, analogous to primary visual processing. 

\section{Code and data}

All code was written in Julia using version 1.12.3; it is available at:\\
\url{github.com/IteratedLM/2025_05_7Seg}\\
The data were generated in Julia and are available at\\
\url{github.com/IteratedLM/2025_12_7Seg_data}
This repository contains Julia and Python scripts for producing the data.

\section{Acknowledgements}

SB was supported by UKRI EPSRC Grant No. EP/Y028392/1: \emph{AI for Collective Intelligence (AI4CI)}. Simulations were carried out using the computational facilities of the Advanced Computing Research Centre at the University of Bristol, \texttt{www.bristol.ac.uk/acrc/}. We are grateful to Dr Stewart whose philanthropy supported some of the compute resource used in this project.

\end{document}